\documentclass[twocolumn, switch]{article}
\usepackage[utf8x]{inputenc}
\usepackage{cite}
\usepackage{amsmath,amssymb,amsfonts}
\usepackage{algorithmic}
\usepackage{graphicx}
\usepackage{adjustbox}
\usepackage{color, colortbl}
\usepackage{booktabs}
\usepackage{textcomp}
\usepackage{url}
\usepackage{multirow}
\usepackage{lipsum}  
\usepackage{bbding}
\usepackage{nicefrac}
\usepackage{preprint}
\usepackage{hyperref}
\usepackage{caption}
\usepackage[numbers,square,sort&compress]{natbib}

\hypersetup{colorlinks=true, linkcolor=purple, urlcolor=blue, citecolor=cyan, anchorcolor=black}

\usepackage{titlesec}
\titlespacing\section{0pt}{12pt plus 3pt minus 3pt}{1pt plus 1pt minus 1pt}
\titlespacing\subsection{0pt}{10pt plus 3pt minus 3pt}{1pt plus 1pt minus 1pt}
\titlespacing\subsubsection{0pt}{8pt plus 3pt minus 3pt}{1pt plus 1pt minus 1pt}

\definecolor{lime}{HTML}{A6CE39}

\title{DiNO-Diffusion: Scaling Medical Diffusion via Self-Supervised Pre-Training}

\author{
Guillermo Jimenez-Perez, Pedro Osorio, Josef Cersovsky, Javier Montalt-Tordera, \\
\textbf{Jens Hooge, Steffen Vogler \& Sadegh Mohammadi} \\
DT\&IT Pharma Decision Science - Computer Vision \& Sound Analysis \\
Diagnostic Imaging Data \& AI \\
Bayer AG \\
\texttt{guillermo@jimenezperez.com, sadegh.mohammadi@bayer.com} \\
}

\newlength{\Oldarrayrulewidth}
\newcommand{\Cline}[2]{%
  \noalign{\global\setlength{\Oldarrayrulewidth}{\arrayrulewidth}}%
  \noalign{\global\setlength{\arrayrulewidth}{#1}}\cline{#2}%
  \noalign{\global\setlength{\arrayrulewidth}{\Oldarrayrulewidth}}}

\begin{document}

\twocolumn[\begin{@twocolumnfalse}

\maketitle

\begin{abstract}
{\small
Diffusion models (DMs) have emerged as powerful foundation models for a variety of tasks, with a large focus in synthetic image generation. However, their requirement of large annotated datasets for training limits their applicability in medical imaging, where datasets are typically smaller and sparsely annotated. We introduce DiNO-Diffusion, a self-supervised method for training latent diffusion models (LDMs) that conditions the generation process on image embeddings extracted from DiNO. By eliminating the reliance on annotations, our training leverages over 868k unlabelled images from public chest X-Ray (CXR) datasets. Despite being self-supervised, DiNO-Diffusion shows comprehensive manifold coverage, with FID scores as low as 4.7, and emerging properties when evaluated in downstream tasks. It can be used to generate semantically-diverse synthetic datasets even from small data pools, demonstrating up to 20\% AUC increase in classification performance when used for data augmentation. Images were generated with different sampling strategies over the DiNO embedding manifold and using real images as a starting point. Results suggest, DiNO-Diffusion could facilitate the creation of large datasets for flexible training of downstream AI models from limited amount of real data, while also holding potential for privacy preservation. Additionally, DiNO-Diffusion demonstrates zero-shot segmentation performance of up to 84.4\% Dice score when evaluating lung lobe segmentation. This evidences good CXR image-anatomy alignment, akin to segmenting using textual descriptors on vanilla DMs. Finally, DiNO-Diffusion can be easily adapted to other medical imaging modalities or state-of-the-art diffusion models, opening the door for large-scale, multi-domain image generation pipelines for medical imaging.
}
\end{abstract}


\vspace{0.5cm}

\end{@twocolumnfalse}]

\section{Introduction}
\label{sec:introduction}

Diffusion models (DMs) have recently emerged as robust and proficient foundational models in medical imaging, exhibiting substantial capabilities in image generation, image enhancement, reconstruction, and segmentation \cite{kazerouni2023medicaldiffusionsurvey}.
The field of synthetic image generation in particular has greatly shifted to text-to-image DMs, generating images that are nearly indistinguishable from real ones \cite{osorio2023latent, chambon2022roentgen, ye2023histodiffusion, aversa2023diffinfinite, pinaya2022brainmridiffusion} and facilitating remarkable zero-shot performance in segmentation and classification tasks \cite{tian2023diffuseattendandsegment, zhang2023ataleoftwofeatures}. However, DMs depend on the availability of large datasets containing images paired with corresponding descriptors (usually text) to guide the generation process, a requirement that presents a considerable obstacle in the medical domain \cite{beddiar2023reviewautomaticcaptioning}. Medical imaging datasets are typically small, contain free-form and inconsistent annotations including captions, binary labels or segmentations, and are generally prohibitively costly to compile and curate \cite{beddiar2023reviewautomaticcaptioning}.

To address these challenges, some works have proposed pseudo-labeling with automatic captioners and other GPT-like vision-language models (VLMs) \cite{betker2023dalle3} or have trained lean mapping networks over frozen pretrained backbones to reduce the number of required annotated samples \cite{li2023blip2, zhang2023controlnet}. However, despite their promise, pseudo-labelling approaches find limited applicability in the medical field given a lack of high-quality medical imaging captioners \cite{beddiar2023reviewautomaticcaptioning}. In addition, while some authors have successfully trained mapping networks to bridge the gap between unimodal foundation models, they still require relatively large annotated datasets to be trained \cite{beddiar2023reviewautomaticcaptioning}.

These limitations are the main roadblocks for medical DMs. While the natural imaging literature focuses on saturating generation quality by improving the base architecture, optimization process or condition alignment \cite{esser2024stablediffusion3, betker2023dalle3, liu2024sora}, the medical imaging community navigates these hurdles by leveraging smaller or custom-annotated datasets \cite{chambon2022roentgen, ye2023histodiffusion, osorio2023latent, aversa2023diffinfinite, pinaya2022brainmridiffusion}. Moreover, although mapping networks have found their footing in the diffusion literature with approaches such as ControlNet \cite{zhang2023controlnet}, these would still rely on large-scale medical DMs trained with prohibitively extensive amounts of annotated images. In this context, applying a self-supervised approach to DM training would be highly beneficial for medical image synthesis. Self-supervision enables models to learn from unlabelled data, providing exceptional results in multiple downstream tasks when used as image embedders \cite{caron2021dino, oquab2023dinov2, perez2024raddino, dippel2024rudolfv, moutakanni2024advancingraydino}.

\begin{figure*}[t]
    \centering
    \includegraphics[width=0.9\textwidth]{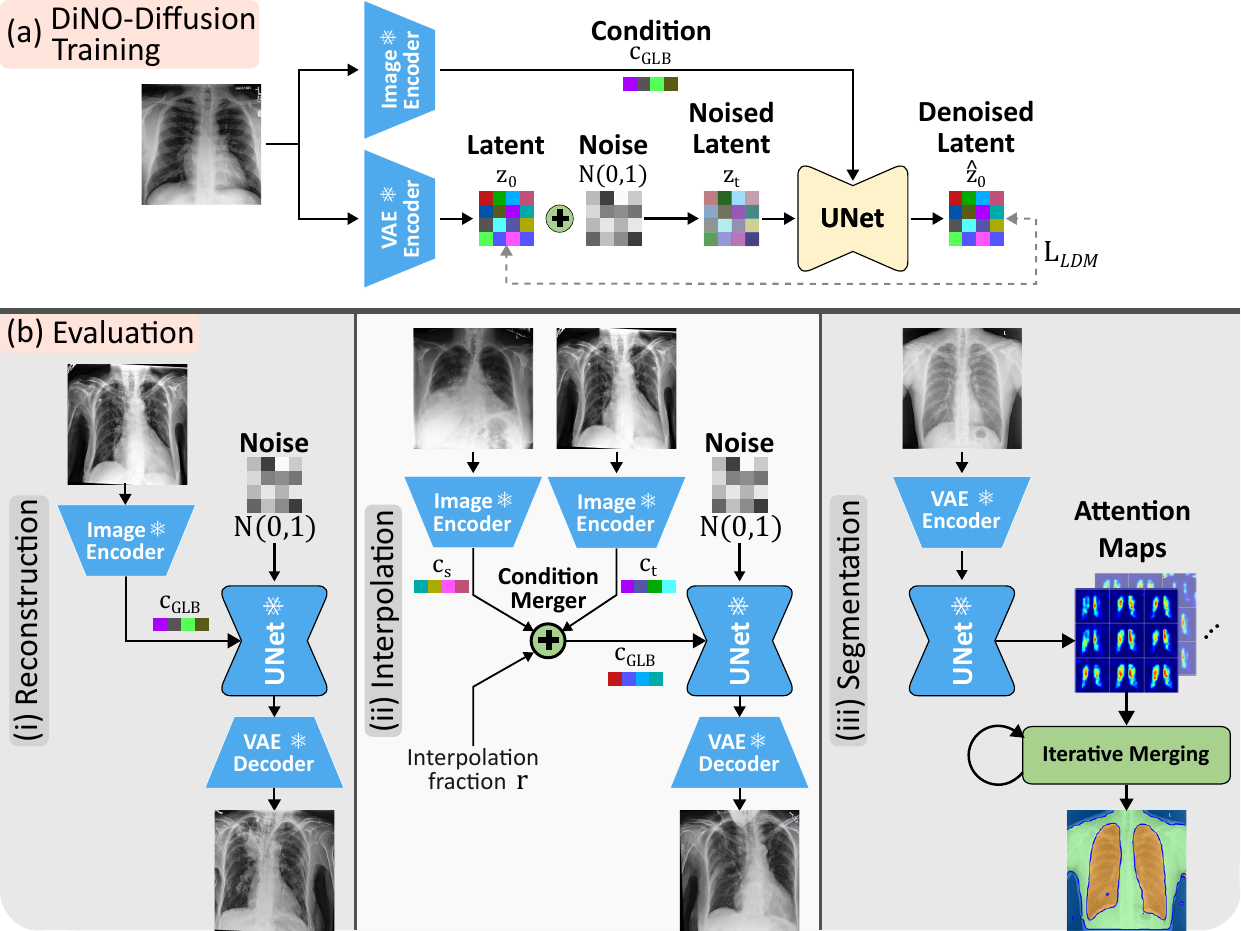}
    \caption{\footnotesize Training and evaluation protocol. (a) DiNO-Diffusion training pipeline: the training image is both embedded into latents $z_0$ with a frozen (\Snowflake) VAE, and  processed by a frozen image encoder to generate global tokens that act as condition $c_{GLB}$. Then, the latents are noised at timestep $z_t$ and fed along the condition to the UNet, which denoises the latent $\hat{z}_0$. Finally, the loss $\textrm{L}_{LDM}$ is computed between $z_0$ and $\hat{z}_0$. (b) Evaluation protocols: the trained UNet is used to produce: (b-i) ``reconstructions'' of a given image; (b-ii) ``interpolated'' synthetic images from the embeddings of a source ($c_s$) and a target ($c_t$) real images at interpolation fraction $r$; or (b-iii) segmentation masks, by iteratively merging latent attention maps.}
    \label{fig:pipeline}
\end{figure*}

With that in mind, we introduce DiNO-Diffusion, a novel self-supervised methodology for training medical DMs at scale which conditions the image generation process on image-derived tokens extracted from a frozen DiNO model \cite{caron2021dino, oquab2023dinov2}, as opposed to textual descriptors. DiNO-Diffusion allows independence from existing annotations, circumventing the limitations imposed by the scarcity and inconsistency of medical image labels. Moreover, it is agnostic to the choice of DM architecture, medical imaging modality or optimization strategy. To test this, a model was trained on a large corpus of open-source CXR data found in the literature \cite{bimcv, chexpert, physionet, indianauniversity, covidgr, montgomeryshenzen, montgomery2, padchest, ralodataset, brax, jrst, kagglepneumoniakermany, covid19imagedatacollection, covid19pakistan1, covid19pakistan2, objectcxr, vindr, vindrpediatric, siim, tbx11k, drugresistanttuberculosis, imagingdatacommons, imagingdatacommonsdata, mimic}, which do not contain any common descriptor to train a regular DM (e.g., text captions). The trained models achieved low FID scores and high image quality. 

DiNO-Diffusion can generate medical images despite using DiNO embeddings, which are derived from natural images. To test the alignment between DiNO embeddings and generated images, three downstream evaluation tasks were performed. Firstly, the potential of the generated images for data augmentation was explored by training a multi-label classifier on increasingly large proportions of synthetic data in addition to the classifier's real training data pool. Secondly, another classifier was trained with synthetic images only, to gauge the extent to which real training data can be replaced with synthetic images. Finally, an image segmentation task was designed to assess whether it is possible to leverage the intermediate image representations learned by the model for creating zero-shot segmentation masks for the distinct anatomical structures.

In summary, our main findings are as follows:
\begin{itemize}
    \item DiNO-Diffusion allows training large DMs given its independence from specific architectures, imaging modalities, available annotations, dataset sizes or optimization strategies. 
    \item DiNO, despite not being trained on medical images, produces embeddings that are descriptive enough for image generation. Using only the global tokens from the embeddings seems to restrict information enough to force DiNO-Diffusion to introduce semantic variability during generation, thus avoiding replication of the input data.
    \item DiNO-Diffusion was used to generate semantically-diverse synthetic datasets even from small data pools. 
    These samples were used for data augmentation, improving classification performance on different data regimes. In addition, training on only synthetic data showed potential for mitigating privacy concerns.
    \item DiNO-Diffusion
    can be leveraged for zero-shot medical image segmentation through iterative attention map merging. This demonstrates its ability to learn semantic coherence and its good alignment with anatomic structures. To our knowledge, this is the first application of zero-shot segmentation applied to medical DMs.
\end{itemize}

\begin{figure*}[t]
    \centering
    \includegraphics[width=0.95\textwidth]{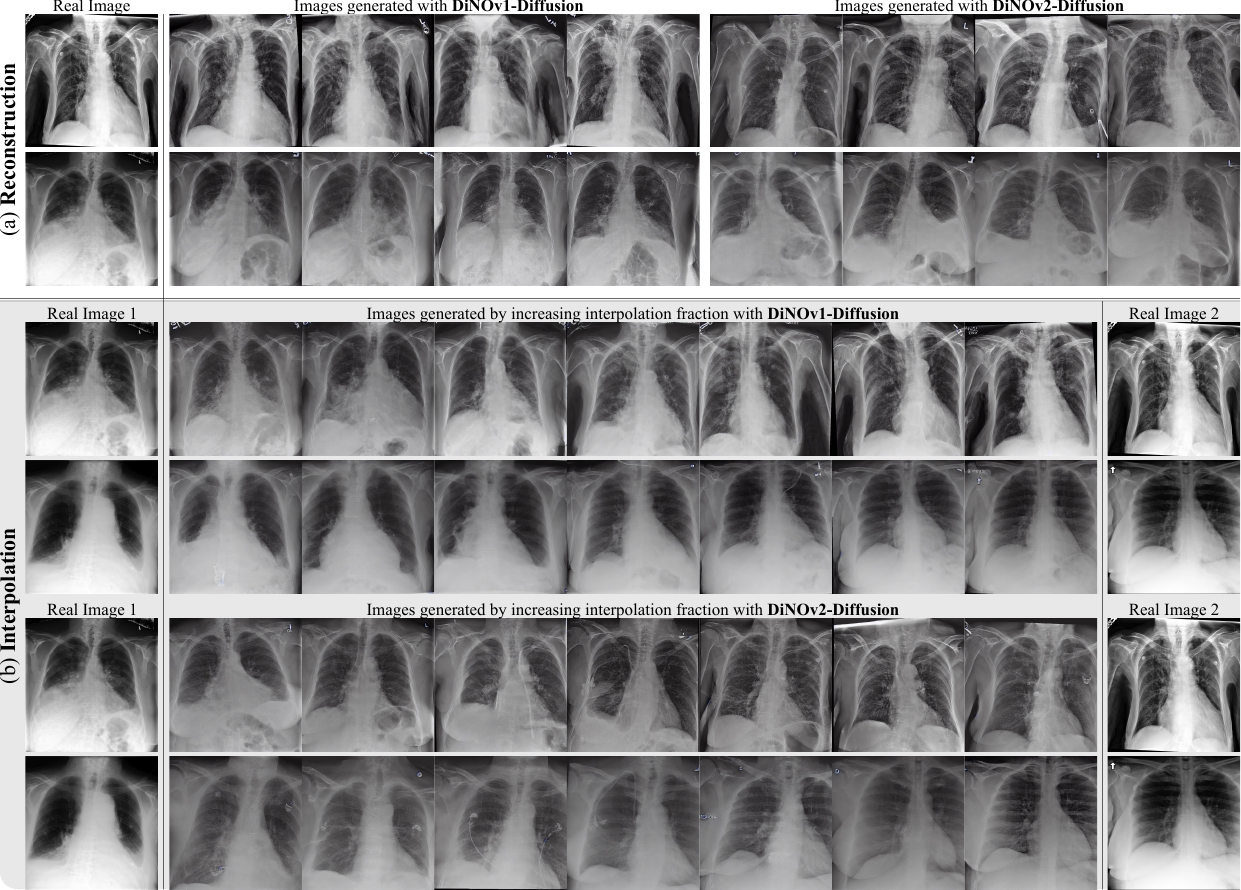}
    \caption{\footnotesize Examples of generated images with DiNO-Diffusion. In the reconstruction experiment (a), each row represents randomly generated examples from two base images within MIMIC and for both DiNOv1-Diffusion and DiNOv2-Diffusion, showing semantic variability. In the interpolation experiment (b), each row depicts two real images and the result from generating synthetic images by interpolating the embeddings incrementally for the DiNOv1-Diffusion (b-top) and DiNOv2-Diffusion (b-bottom) settings.}
    \label{fig:results-generated-images}
\end{figure*}

\section{Methods}
\label{sec:methods}

This Section explains the methodology employed for studying the self-supervised DM. In Section \ref{sec:materials}, the datasets used for training and evaluation are described. In Section \ref{sec:methods-architecture}, the model's architecture and theoretical background is outlined. In Section \ref{sec:methods-conditioning}, the designed mechanisms for self-supervised conditioning are detailed. In Section \ref{sec:methods-evaluation}, the evaluation protocol employed to benchmark model performance is defined. Finally, in Section \ref{sec:methods-experimental-setup}, the specific parameters used for model training and evaluation are enumerated. Figure~\ref{fig:pipeline} visually describes the training and evaluation pipeline.

\subsection{Data}
\label{sec:materials}

To explore DiNO-Diffusion's self-supervision capability, a large-scale dataset comprised of every
openly accessible CXR dataset found in the literature \cite{bimcv, chexpert, physionet, mimic, indianauniversity, covidgr, montgomeryshenzen, montgomery2, padchest, ralodataset, brax, jrst, kagglepneumoniakermany, covid19imagedatacollection, covid19pakistan1, covid19pakistan2, objectcxr, vindr, vindrpediatric, siim, tbx11k, drugresistanttuberculosis, imagingdatacommons, imagingdatacommonsdata}\footnote{Thanks, among others, to the National Library of Medicine, National Institutes of Health, Bethesda, MD, USA.} was collected, reaching over 1.2M total images from 21 distinct data providers. 
Three different subsets were taken from this compound dataset for different purposes. Firstly, a subset comprising every dataset minus MIMIC-CXR \cite{mimic} was selected for training the DiNO-Diffusion models. Their labels were discarded and label balancing was not performed, resulting in 868 394 samples with a variety of image sources, resolutions and patient characteristics. Secondly, MIMIC-CXR was used solely for evaluating the model via two classification tasks (see Section~\ref{sec:methods-evaluation}). MIMIC-CXR is composed of chest radiographs with free-text radiology reports, for which multi-label classification information is available. The MIMIC-CXR dataset was preprocessed to match similar literature \cite{chambon2022roentgen, ktena2024nature} by discarding lateral views, by restricting the labels to those whose prevalence was of at least 4\% (Atelectasis, Cardiomegaly, Consolidation, Edema, Pleural  Effusion, Pneumonia and Pneumothorax), and by splitting its p10-p18 subsets for classifier training and leaving p19 as a held-out test set. Finally, the third subset for the segmentation task relied on three small datasets containing annotated masks: the JSRT ($N=247$), Montgomery ($N=138$) and Shenzhen ($N=663$) \cite{jrst, montgomeryshenzen} datasets.

\subsection{Generative Architecture - Stable Diffusion}
\label{sec:methods-architecture}

Latent Diffusion Models (LDMs) approach image generation as an iterative denoising process, transforming pure noise $x_T$ into a defined image $x_0$ over $T$ steps with a parameterized DM $\epsilon_\theta(z_t, t, c)$, where $c$ represents an optional condition. LDMs address the prohibitive computational demands of traditional DMs by reducing the dimensionality of the input. LDMs currently find active development with ongoing research in different parameterised models, optimization strategies and dimensionality reduction pipelines.

This study adopts the Stable Diffusion (SD) framework (version 1) \cite{rombach2022stablediffusion} as its baseline. Despite being outperformed by more recent models and its output size limitation of 512x512 pixels, SD's lightweight architecture, open-source nature, and community adoption makes it ideal for our proof of concept. SD comprises a frozen \textit{variational autoencoder} (VAE) and a trainable \textit{conditional denoising UNet}.

The VAE consists of an encoder ($\mathcal{E}$) and a decoder  ($\mathcal{D}$). The encoder compresses fixed-size images $x \in \mathbb{R}^{H \times W \times 3}$ into a latent $z = \mathcal{E}(x) \in \mathbb{R}^{(H/d) \times (W/d) \times k}$, where $k = 4$ is number of channels extracted by the VAE and $d = 8$ is 
the downsampling factor. The decoder maps latents back to the original image space $\hat{x} = \mathcal{D}(z)$. Stable Diffusion's VAE has been shown to generalize to medical data \cite{chambon2022roentgen, chambon2022adapting}. The UNet serves as the diffusion component and uses a ResNet architecture as its convolutional backbone, where the condition $c$ is incorporated through attention mechanisms (see Section~\ref{sec:methods-conditioning}). 

With this model, training with conditional information involves two phases: the \textit{forward} and \textit{reverse diffusion} processes. During the \textit{forward} diffusion, an image $x_0$ (or its latent representation $z_0$) and condition $c$ are chosen. A timestep $t$ is randomly selected ($t \sim \mathcal{U}({1, ..., T})$) so a noisy latent $z_t$ is generated by mixing $z_0$ with noise $\epsilon \sim \mathcal{N}(0,1)$, resulting in a \textit{partially noised} latent. The \textit{reverse} process uses the UNet to estimate the original noise $\epsilon$ from $z_t$, $t$ and $c$. 

The network is optimized using the Mean Squared Error (MSE) loss between the predicted and actual noise to adjust the weights of the UNet:
\begin{equation}
    \textrm{L}_{LDM} = \textrm{E}_{z \sim \epsilon(x), \, c, \, \epsilon \sim \mathcal{N} (0, 1), \, t} \left[ {|| \epsilon - \epsilon_\theta(z_t, t, c)||}_2^2 \right]
\end{equation}
After training, image synthesis begins with sampling a noisy latent $z_T \sim \mathcal{N}(0,1)$, progressively denoising it with condition $c$ to obtain $z_0$ so that $\hat{z}_0 = \epsilon_\theta(z_{T:0}, c)$, and by using the VAE's decoder, so that $\hat{x} = \mathcal{D}(\hat{z}_0) = \mathcal{D}(\epsilon_\theta(z_{T:0}, c))$.

\subsection{Self-Supervised Conditioning}
\label{sec:methods-conditioning}

LDMs condition image generation using a semantic tensor $c$ to guide the diffusion process. This tensor is usually obtained from a frozen transformer model $f_\Phi$ that maps the label information into a tensor $c = f_\Phi(x) \in \mathcal{R}^{S \times N}$, where $S$ is the token length (of variable size), $N$ is the embedding dimension and $x$ represents whichever input the embedder model requires (text, image, etc.). Although the current diffusion literature has mainly focused on using textual descriptors as their main conditioning strategy, other conditioning mechanisms have been employed \cite{aversa2023diffinfinite, pinaya2022brainmridiffusion, zhang2023controlnet}. 

\begin{table*}[t]
\centering
\begin{adjustbox}{width=1\textwidth}
\begin{tabular}{|lll|c|c|c|c|c|c|}
\hline
\rowcolor[gray]{.85}
\multicolumn{3}{|c|}{Strategy}                                       & 
$rs$ ratio
& $AUC_{N = 50} \downarrow$ & $AUC_{N = 100} \downarrow$ & $AUC_{N = 500} \downarrow$ & $AUC_{N = 1000} \downarrow$ & $AUC_{N = 5000} \downarrow$ \\ \hline \hline 
\rowcolor[RGB]{223,230,245}
\multicolumn{3}{|c|}{Real data} & 1:0 (real-only)        & $0.548 \pm 0.013$                & $0.566 \pm 0.047$                 & $0.682 \pm 0.011$                 & $0.715 \pm 0.005$                  & $0.747 \pm 0.006$ \\ \hline \hline
\cellcolor[RGB]{247,243,212} & \multicolumn{1}{|l|}{}                                    &                                  & 1:1                   & $0.551 \pm 0.037$                & $0.602 \pm 0.025$                 & $0.685 \pm 0.012$                 & $\boldsymbol{0.724 \pm 0.002}$ *                 & $0.756 \pm 0.002$ *                 \\ \cline{4-9} 
\cellcolor[RGB]{247,243,212} & \multicolumn{1}{|l|}{}                                    &                                  & 1:5                   & $0.564 \pm 0.050$                & $0.626 \pm 0.016$                 & $\boldsymbol{0.706 \pm 0.010}$ *                 & $0.725 \pm 0.005$                  & $\boldsymbol{0.756 \pm 0.003}$ *                  \\ \cline{4-9} 
\cellcolor[RGB]{247,243,212} & \multicolumn{1}{|l|}{}                                    &                                  & 1:10                  & $0.608 \pm 0.024$ *                & $0.618 \pm 0.030$                 & $0.701 \pm 0.014$                 & $0.719 \pm 0.007$                  & $0.745 \pm 0.012$                  \\ \cline{4-9} 
\cellcolor[RGB]{247,243,212} & \multicolumn{1}{|l|}{}                                    & \multirow{-4}{*}{\rotatebox[origin=c]{90}{\begin{tabular}[c]{@{}c@{}}Recons-\\ truction\end{tabular}}} & 1:50                  & $\boldsymbol{0.650 \pm 0.020}$ *                & $\boldsymbol{0.651 \pm 0.013}$ *                 & $0.698 \pm 0.009$                 & $0.699 \pm 0.012$                  & $0.735 \pm 0.006$                  \\ \Cline{1.00pt}{3-9} 
\cellcolor[RGB]{247,243,212} & \multicolumn{1}{|l|}{}                                    &                                  & 1:1        & $0.540 \pm 0.036$                & $0.589 \pm 0.033$                 & $0.676 \pm 0.007$                 & $0.682 \pm 0.011$ *                 & $0.686 \pm 0.009$ *                 \\ \cline{4-9} 
\cellcolor[RGB]{247,243,212} & \multicolumn{1}{|l|}{}                                    &                                  & 1:5                   & $0.579 \pm 0.033$                & $0.625 \pm 0.011$                 & $0.696 \pm 0.013$                 & $0.706 \pm 0.007$ *                 & $0.703 \pm 0.007$ *                 \\ \cline{4-9} 
\cellcolor[RGB]{247,243,212} & \multicolumn{1}{|l|}{}                                    &                                  & 1:10                   & $0.589 \pm 0.039$ *                & $0.618 \pm 0.018$                 & $\boldsymbol{0.709 \pm 0.009}$ *                 & $0.709 \pm 0.003$                  & $0.693 \pm 0.018$ *                  \\ \cline{4-9} 
\cellcolor[RGB]{247,243,212} & \multicolumn{1}{|l|}{\multirow{-8}{*}{\rotatebox[origin=c]{90}{DiNOv1-Diffusion}}} & \multirow{-4}{*}{\rotatebox[origin=c]{90}{\begin{tabular}[c]{@{}c@{}}Inter-\\ polation\end{tabular}}}  & 1:50                  & $\boldsymbol{0.632 \pm 0.015}$ *                & $\boldsymbol{0.644 \pm 0.014}$ *                 & $0.702 \pm 0.013$                 & $\boldsymbol{0.716 \pm 0.013}$                  & $0.743 \pm 0.004$                  \\ \cline{2-9} \cline{2-9}
\cellcolor[RGB]{247,243,212} & \multicolumn{1}{|l|}{}                                    &                                  & 1:1                   & $0.515 \pm 0.026$                & $0.566 \pm 0.015$                 & $0.692 \pm 0.022$                 & $0.716 \pm 0.008$                  & $\boldsymbol{0.747 \pm 0.003}$                  \\ \cline{4-9} 
\cellcolor[RGB]{247,243,212} & \multicolumn{1}{|l|}{}                                    &                                  & 1:5                   & $0.552 \pm 0.036$                & $0.608 \pm 0.035$                 & $\boldsymbol{0.705 \pm 0.004}$ *                 & $0.714 \pm 0.006$                  & $0.744 \pm 0.004$                  \\ \cline{4-9} 
\cellcolor[RGB]{247,243,212} & \multicolumn{1}{|l|}{}                                    &                                  & 1:10                  & $0.611 \pm 0.010$ *                & $0.631 \pm 0.029$                 & $0.705 \pm 0.006$ *                 & $\boldsymbol{0.717 \pm 0.005}$                  & $0.745 \pm 0.006$                  \\ \cline{4-9} 
\cellcolor[RGB]{247,243,212} & \multicolumn{1}{|l|}{}                                    & \multirow{-4}{*}{\rotatebox[origin=c]{90}{\begin{tabular}[c]{@{}c@{}}Recons-\\ truction\end{tabular}}} & 1:50                  & $\boldsymbol{0.617 \pm 0.018}$ *                & $\boldsymbol{0.627 \pm 0.016}$ *                 & $0.700 \pm 0.016$                 & $0.710 \pm 0.005$                  & $0.744 \pm 0.004$                  \\ \Cline{1.00pt}{3-9} 
\cellcolor[RGB]{247,243,212} & \multicolumn{1}{|l|}{}                                    &                                  & 1:1        & $0.574 \pm 0.043$                & $0.603 \pm 0.049$                 & $\boldsymbol{0.685 \pm 0.009}$                 & $0.698 \pm 0.007$ *                 & $0.681 \pm 0.011$ *                 \\ \cline{4-9} 
\cellcolor[RGB]{247,243,212} & \multicolumn{1}{|l|}{}                                    &                                  & 1:5                   & $0.580 \pm 0.018$ *                & $0.594 \pm 0.053$                 & $0.657 \pm 0.023$                 & $0.688 \pm 0.011$ *                 & $0.710 \pm 0.008$ *                 \\ \cline{4-9} 
\cellcolor[RGB]{247,243,212} & \multicolumn{1}{|l|}{}                                    &                                  & 1:10                   & $0.608 \pm 0.025$ *                & $0.622 \pm 0.026$                 & $0.681 \pm 0.017$                 & $0.694 \pm 0.005$ *                 & $0.689 \pm 0.021$ *                  \\ \cline{4-9} 
\multirow{-16}{*}{\rotatebox[origin=c]{90}{\cellcolor[RGB]{247,243,212} \textbf{(a) Data Augmentation}}} & \multicolumn{1}{|l|}{\multirow{-8}{*}{\rotatebox[origin=c]{90}{DiNOv2-Diffusion}}} & \multirow{-4}{*}{\rotatebox[origin=c]{90}{\begin{tabular}[c]{@{}c@{}}Inter-\\ polation\end{tabular}}}  & 1:50                  & $\boldsymbol{0.618 \pm 0.020}$ *                & $\boldsymbol{0.649 \pm 0.016}$ *                 & $0.690 \pm 0.024$                 & $0.703 \pm 0.008$ *                  & $0.702 \pm 0.013$ *                  \\ \hline 
    \multicolumn{9}{c}{} \\ 
\hline 
\cellcolor[RGB]{229,248,229} & \multicolumn{1}{|l|}{}                                    &                                  & 1:1                  & $0.546 \pm 0.017$                & $0.571 \pm 0.046$                 & $0.667 \pm 0.008$ *                 & $0.696 \pm 0.010$ *                  & $0.730 \pm 0.004$ *                  \\ \cline{4-9} 
\cellcolor[RGB]{229,248,229} & \multicolumn{1}{|l|}{}                                    &                                  & 1:5                  & $0.574 \pm 0.059$                & $0.610 \pm 0.029$ *                 & $\boldsymbol{0.701 \pm 0.007}$                  & $\boldsymbol{0.724 \pm 0.004}$ *                  & $0.752 \pm 0.005$                  \\ \cline{4-9} 
\cellcolor[RGB]{229,248,229} & \multicolumn{1}{|l|}{}                                    &                                  & 1:10                 & $0.625 \pm 0.020$ *                & $0.631 \pm 0.025$ *                 & $0.701 \pm 0.010$                 & $0.722 \pm 0.005$                  & $\boldsymbol{0.753 \pm 0.006}$                  \\ \cline{4-9} 
\cellcolor[RGB]{229,248,229} & \multicolumn{1}{|l|}{}                                    & \multirow{-4}{*}{\rotatebox[origin=c]{90}{\begin{tabular}[c]{@{}c@{}}Recons-\\ truction\end{tabular}}} & 1:50                 & $\boldsymbol{0.655 \pm 0.015}$ *                & $\boldsymbol{0.645 \pm 0.011}$ *                 & $0.689 \pm 0.018$                 & $0.709 \pm 0.014$                  & $0.746 \pm 0.006$                  \\ \Cline{1.00pt}{3-9} 
\cellcolor[RGB]{229,248,229} & \multicolumn{1}{|l|}{}                                    &                                  & 1:1       & $0.515 \pm 0.029$                & $0.491 \pm 0.033$                 & $0.530 \pm 0.035$ *                 & $0.546 \pm 0.016$ *                  & $0.538 \pm 0.020$ *                  \\ \cline{4-9} 
\cellcolor[RGB]{229,248,229} & \multicolumn{1}{|l|}{}                                    &                                  & 1:5                  & $0.525 \pm 0.015$ *                & $0.576 \pm 0.037$                 & $0.686 \pm 0.011$                  & $0.695 \pm 0.008$ *                  & $0.531 \pm 0.009$ *                  \\ \cline{4-9} 
\cellcolor[RGB]{229,248,229} & \multicolumn{1}{|l|}{}                                    &                                  & 1:10                  & $0.572 \pm 0.023$ *                & $0.574 \pm 0.013$                 & $0.701 \pm 0.005$ *                 & $0.706 \pm 0.005$ *                  & $0.686 \pm 0.005$ *                  \\ \cline{4-9} 
\cellcolor[RGB]{229,248,229} & \multicolumn{1}{|l|}{\multirow{-8}{*}{\rotatebox[origin=c]{90}{DiNOv1-Diffusion}}} & \multirow{-4}{*}{\rotatebox[origin=c]{90}{\begin{tabular}[c]{@{}c@{}}Inter-\\ polation\end{tabular}}}  & 1:50                 & $\boldsymbol{0.635 \pm 0.018}$ *                & $\boldsymbol{0.644 \pm 0.015}$ *                 & $\boldsymbol{0.705 \pm 0.013}$ *                 & $0.711 \pm 0.011$                  & $0.736 \pm 0.007$                  \\ \cline{2-9} \cline{2-9}
\cellcolor[RGB]{229,248,229} & \multicolumn{1}{|l|}{}                                    &                                  & 1:1                  & $0.509 \pm 0.025$ *                & $0.564 \pm 0.044$                 & $0.646 \pm 0.021$ *                 & $0.649 \pm 0.005$ *                  & $0.711 \pm 0.004$ *                  \\ \cline{4-9} 
\cellcolor[RGB]{229,248,229} & \multicolumn{1}{|l|}{}                                    &                                  & 1:5                  & $0.523 \pm 0.019$ *                & $0.591 \pm 0.048$                 & $0.684 \pm 0.009$                  & $0.700 \pm 0.007$ *                  & $0.728 \pm 0.007$ *                  \\ \cline{4-9} 
\cellcolor[RGB]{229,248,229} & \multicolumn{1}{|l|}{}                                    &                                  & 1:10                 & $0.574 \pm 0.034$                & $0.610 \pm 0.021$ *                 & $0.687 \pm 0.012$                 & $0.695 \pm 0.011$ *                  & $0.730 \pm 0.006$ *                  \\ \cline{4-9} 
\cellcolor[RGB]{229,248,229} & \multicolumn{1}{|l|}{}                                    & \multirow{-4}{*}{\rotatebox[origin=c]{90}{\begin{tabular}[c]{@{}c@{}}Recons-\\ truction\end{tabular}}} & 1:50                 & $\boldsymbol{0.603 \pm 0.033}$ *                & $\boldsymbol{0.626 \pm 0.015}$ *                 & $\boldsymbol{0.699 \pm 0.014}$ *                 & $0.708 \pm 0.006$                  & $0.741 \pm 0.006$                  \\ \Cline{1.00pt}{3-9} 
\cellcolor[RGB]{229,248,229} & \multicolumn{1}{|l|}{}                                    &                                  & 1:1       & $0.546 \pm 0.045$                & $0.567 \pm 0.019$                 & $0.553 \pm 0.035$ *                 & $0.558 \pm 0.015$ *                  & $0.533 \pm 0.024$ *                  \\ \cline{4-9} 
\cellcolor[RGB]{229,248,229} & \multicolumn{1}{|l|}{}                                    &                                  & 1:5                  & $0.536 \pm 0.030$                & $0.593 \pm 0.040$                 & $0.631 \pm 0.029$ *                 & $0.669 \pm 0.008$ *                  & $0.646 \pm 0.016$ *                  \\ \cline{4-9} 
\cellcolor[RGB]{229,248,229} & \multicolumn{1}{|l|}{}                                    &                                  & 1:10                  & $0.551 \pm 0.030$                & $0.602 \pm 0.035$                 & $0.668 \pm 0.017$                 & $0.660 \pm 0.014$ *                  & $0.680 \pm 0.017$ *                  \\ \cline{4-9} 
\multirow{-16}{*}{\rotatebox[origin=c]{90}{\cellcolor[RGB]{229,248,229} \textbf{(b) Full Synthetic Training}}} & \multicolumn{1}{|l|}{\multirow{-8}{*}{\rotatebox[origin=c]{90}{DiNOv2-Diffusion}}} & \multirow{-4}{*}{\rotatebox[origin=c]{90}{\begin{tabular}[c]{@{}c@{}}Inter-\\ polation\end{tabular}}}  & 1:50                 & $\boldsymbol{0.610 \pm 0.052}$ *                & $\boldsymbol{0.625 \pm 0.009}$ *                 & $0.672 \pm 0.018$                 & $0.677 \pm 0.005$ *                  & $0.714 \pm 0.016$ *                  \\ \hline
\end{tabular}
\end{adjustbox}
\caption{AUC scores (mean $\pm$ SD; five-fold cross-validation) for (a) data augmentation and (b) full synthetic trainings across DiNO-Diffusion variants, image synthesis strategies (reconstruction, interpolation), different real-to-synthetic ratios ($rs$) and data regimes ($N$). In (a), classification models are trained with both real and synthetic data, where in (b) only synthetic data was employed. Finally, the baseline (i.e., training with real data only) test performances are depicted at the top in light-blue. \textbf{Bold} values represent best and most significant performance improvement relative to the real-only baseline for each data regime, DiNO-Diffusion model and synthesis strategy. Asterisks (*) represent statistical significance ($p < 0.05$).}
\label{tab:results-classification-experiments}
\end{table*}

In this work we explore conditioning using image-derived semantic descriptors. Specifically, a vision transformer trained with the DiNO method \cite{dosovitskiy2020image, caron2021dino} was used to produce a semantic description of the image to be generated. Vision transformers split an image into small patches (usually $P = 14px^2$ or $P = 16px^2$) representing ``visual words'' and operate over them using a standard transformer architecture. The model outputs a tensor of tokens $c = f_\Phi(x) \in \mathcal{R}^{S \times N}$ comprising a class token $c_{CLS} \in \mathcal{R}^{N}$, sometimes a pooler token $c_{PLR}  \in \mathcal{R}^{N}$, sometimes a predefined amount $R$ of register tokens $c_{REG} \in \mathcal{R}^{R \times N}$ \cite{darcet2023registers}, and finally a series of $L$ patch tokens $c_{LCL} \in \mathcal{R}^{L \times N}$, where $L = H/P_y * W/P_x$. Finally, the conditioning tensor outputted by the embedder was reduced to the available global information $c_{GLB} = [c_{CLS}, c_{PLR}, c_{REG}]$ before feeding it to the UNet, as upon initial exploration the patch tokens contained too much local information of the original image $x$ and led to trivial models that learnt to reconstruct images from redundant information. Figure~\ref{fig:pipeline}-(a) visually describes the training pipeline.

Conditioning image generation on image embeddings offers flexibility on generation as long as a conditioning embedding exists. In this work, two simple generation strategies were explored, to evaluate the model's in-distribution and out-of-distribution performance, although more advanced approaches could be devised.

\subsubsection{Reconstruction-based image generation}
\label{sec:methodology-generation-reconstruction}

the ``reconstruction'' strategy consists in synthesizing images $\hat{x} = \mathcal{D}(\epsilon_\theta(z_{T:0}, c))$ from the global information of an existing real example $(x, y)$, where $y$ is the image's label, $\hat{y} = y$ and $f_\Phi(x)$ is the conditioning embedding as produced by DiNO. 
This reconstruction leverages DiNO-Diffusion's large-scale pretraining to produce semantic variations over the source image $x$. Exact replicas of $x$ are prevented by design due to conditioning with the compressed information from DiNO's global embedding, causing a bottleneck.
Figure~\ref{fig:pipeline} (b-i) depicts the reconstruction process.

\subsubsection{Interpolation-based image generation}
\label{sec:methodology-generation-interpolation}

the ``interpolation'' strategy uses the same image generation mechanism from above. The difference lies in the sampling method of the conditioning embedding $c$, which is interpolated from two images $(x_1, y_1), (x_2, y_2)$ so that $\hat{c} = lerp(f_\Phi(x_1), f_\Phi(x_2), r)$, where $r \in [0,1]$ is the interpolation fraction. This strategy attempts to generate synthetic images from less sampled regions of the real data manifold, located between existing samples, following approaches such as MixUp \cite{zhang2017mixup}. See Figure~\ref{fig:pipeline} (b-ii) for a visual depiction of this strategy. 

\subsection{Evaluation}
\label{sec:methods-evaluation}

This section details three different evaluation protocols used for benchmarking DiNO-Diffusion. For every evaluation protocol, two variants are evaluated and compared: DiNOv1-Diffusion \cite{caron2021dino} and DiNOv2-Diffusion \cite{oquab2023dinov2}, depending on the choice of image encoder.

\subsubsection{Image Quality \& Checkpoint Selection}
\label{sec:methods-evaluation-image-quality}

\begin{figure}[t]
    \centering
    \includegraphics[width=\linewidth]{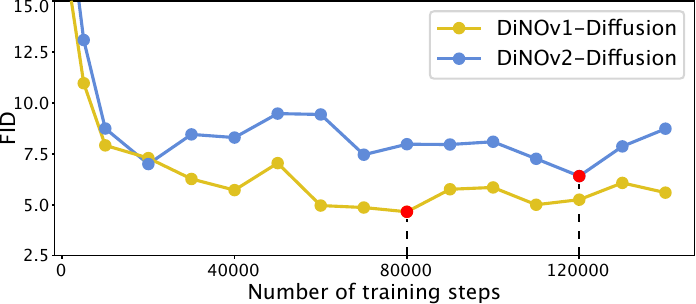}
    \caption{\footnotesize FID scores for both DiNO-Diffusion models, computed every 2500 steps over a subset of MIMIC. Lower is better.}
    \label{fig:imagequalitymetrics}
\end{figure}

Fréchet Inception Distance (FID) \cite{heusel2017fid} was used to quantify generation quality at multiple checkpoints for both variants of DiNO-Diffusion. The FID scores computed for the data generated via the ``reconstruction'' strategy (see Section~\ref{sec:methods-conditioning}) were used as a proxy for overall model performance.
Similarly to \cite{chambon2022roentgen}, FID scores were computed over a 5k subset of MIMIC-CXR’s p19 dataset \cite{mimic}, and are reported every 2500 steps in Figure~\ref{fig:imagequalitymetrics}. Also following \cite{chambon2022roentgen}, the FID score was computed on the feature space of a pretrained domain-specific image encoder from TorchXrayVision \cite{cohen2022torchxrayvision} as opposed to the default Inception-V3 model, as the latter might not provide an accurate measure of image quality when dealing with medical image data. Finally, the optimal checkpoint for each DiNO-Diffusion model was the checkpoint with the lowest FID score.

\subsubsection{Data Augmentation}
\label{sec:methods-evaluation-data-augmentation}

this experiment explored DiNO-Diffusion's ability to enhance the sample size of a dataset by training a classification model on real and synthetic data using five-fold cross-validation and testing on a held-out test set (MIMIC's p19). For this purpose, MIMIC's training dataset (p10-p18) was subset into different data regimes with decreasing sample size $\mathcal{X}_n$, with $n \in \{10k, 5k, 1k, 500, 100, 50\}$ samples in the subset. Given that MIMIC has multi-label annotations, label balancing was performed by randomly selecting $n / card(\mathcal{L})$ elements of each label in the labelset $\mathcal{L}$ from $\mathcal{X}$ without replacement, ensuring sufficient representativity of all labels within the training set. Smaller subsets were also enforced to be contained into bigger ones, so that $\mathcal{X}_{\mathcal{N}_{i+1}} \in \mathcal{X}_{\mathcal{N}_{i}}$. With $\mathcal{X}_n$ defined, synthetic data was created to increase sample size by generating partially-synthetic datasets $\hat{\mathcal{X}}_n$ with real-to-synthetic ratios of 1:1, 1:5, 1:10 and 1:50 for the reconstruction- and interpolation-based synthesis (see Section~\ref{sec:methods-conditioning}).

For the reconstruction experiments, ratios larger than 1:1 represent several semantic variations of a single source image $(x, y)$, with the intent to introduce realistic variance into the synthetic data while retaining the label-specific image features. The interpolation experiment addressed whether intermediate embeddings could still be decoded into an image that retains label-specific features from both elements in the pair. For this purpose, the sample pairs were enforced to have at least one label in common (see Section~\ref{sec:methodology-generation-interpolation}) without repetition. When not all the labels are in common between the pair, the labels of the interpolated example are set to the ones of the sample it is closest to, as defined by the interpolation fraction $r$. Finally, in the case of not having enough unique pairs for a given split, some pairings were repeated with different $r$. 

\subsubsection{Full Synthetic Training}
\label{sec:methods-evaluation-full-synthetic}

this experiment explores whether test-set AUC drops when training a classifier solely on synthetic data, to address whether DiNO-Diffusion can serve as a privacy-preserving synthetic replacement for real data. The generation strategies, data regimes, real-to-synthetic ratios and 5-fold cross-validation settings from Section~\ref{sec:methods-evaluation-data-augmentation} were followed as evaluation strategy.

\subsubsection{Zero-Shot Segmentation}
\label{sec:methods-evaluation-segmentation}

this experiment investigates the model's ability to learn semantic coherence by generating segmentation masks from the internal representations generated during the DiNO-Diffusion's UNet forward pass. For this purpose, the zero-shot segmentation approach from DiffSeg \cite{tian2023diffuseattendandsegment} was followed, consisting of leveraging the self-attention weights from each transformer block of the UNet and iteratively merging them based on their Kullback-Leibler divergence. This methodology was applied both to DiNO-Diffusion and a vanilla SD model to generate lung lobe segmentation masks without further training. 
Using a combined dataset of 1,048 cases with ground truth annotations (See Section~\ref{sec:materials}), candidate masks were evaluated by their Dice score and selected via non-maximum suppression. The respective hyperparameters (merging threshold, timestep, number of anchor points) as well as the best performing checkpoint were selected per model using grid-search. Refer to Figure~\ref{fig:pipeline} (b-iii) for a visual depiction of the segmentation pipeline.

\subsection{Experimental Setup}
\label{sec:methods-experimental-setup}

The models were trained by adapting HuggingFace Diffusers' script for training DMs \cite{von-platen-etal-2022-diffusers-text2im}. The DMs were trained for 100 epochs ($\sim140000$ steps) using 4 H100 GPUs per model, an aggregated batch size of 512 ($bs = 64$, gradient accumulation of 2 steps), 8-bit Adam optimizer with constant $lr = 10^{-4}$ and 1000-step warmup and xformers' memory-efficient attention \cite{facebook2022xformers}. 

The specific versions of the DiNOv1 and DiNOv2 image encoder architectures used were ``\texttt{facebook/dino-vitb16}'' \cite{caron2021dino} and ``\texttt{timm/vit-base-patch14-reg4-dinov2}'' \cite{oquab2023dinov2, darcet2023registers}, respectively. The webdataset \cite{webdataset} library was used for storing data and to stream it directly from the bucket during all model trainings. The classification experiments were based on training HuggingFace's implementation of a ``\texttt{densenet121}'' for 150 max epochs using T4 GPUs with batch size 64, AdamW optimizer with $lr = 10^{-4}$ and weight decay of $10^{-5}$, a LR reduction-on-plateau scheduler with patience 10 and early stopping after 25 epochs with no validation AUC improvement. For the checkpoint evaluation, the specific version of the feature extractor was ``\texttt{densenet121-res224-all}'', a pretrained DenseNet-121 on CXR data from \cite{cohen2022torchxrayvision}.

All images followed the same minimal preprocessing strategy before training or evaluation, similar to other works in the literature \cite{cohen2022torchxrayvision, chambon2022roentgen}. Dynamic intensity values (uint8, uint16) were rescaled to uint8. Images were center-cropped with a 1:1 aspect ratio, resized to 512x512 pixels and padded areas were removed. Minimal data augmentations were applied during all model trainings, including random sharpening and affine transformations (5\% shearing, 5\% translation, 90\%-140\% scaling).

\section{Results}
\label{sec:results}

\begin{figure*}[!t]
    \centering
    \includegraphics[width=0.95\linewidth]{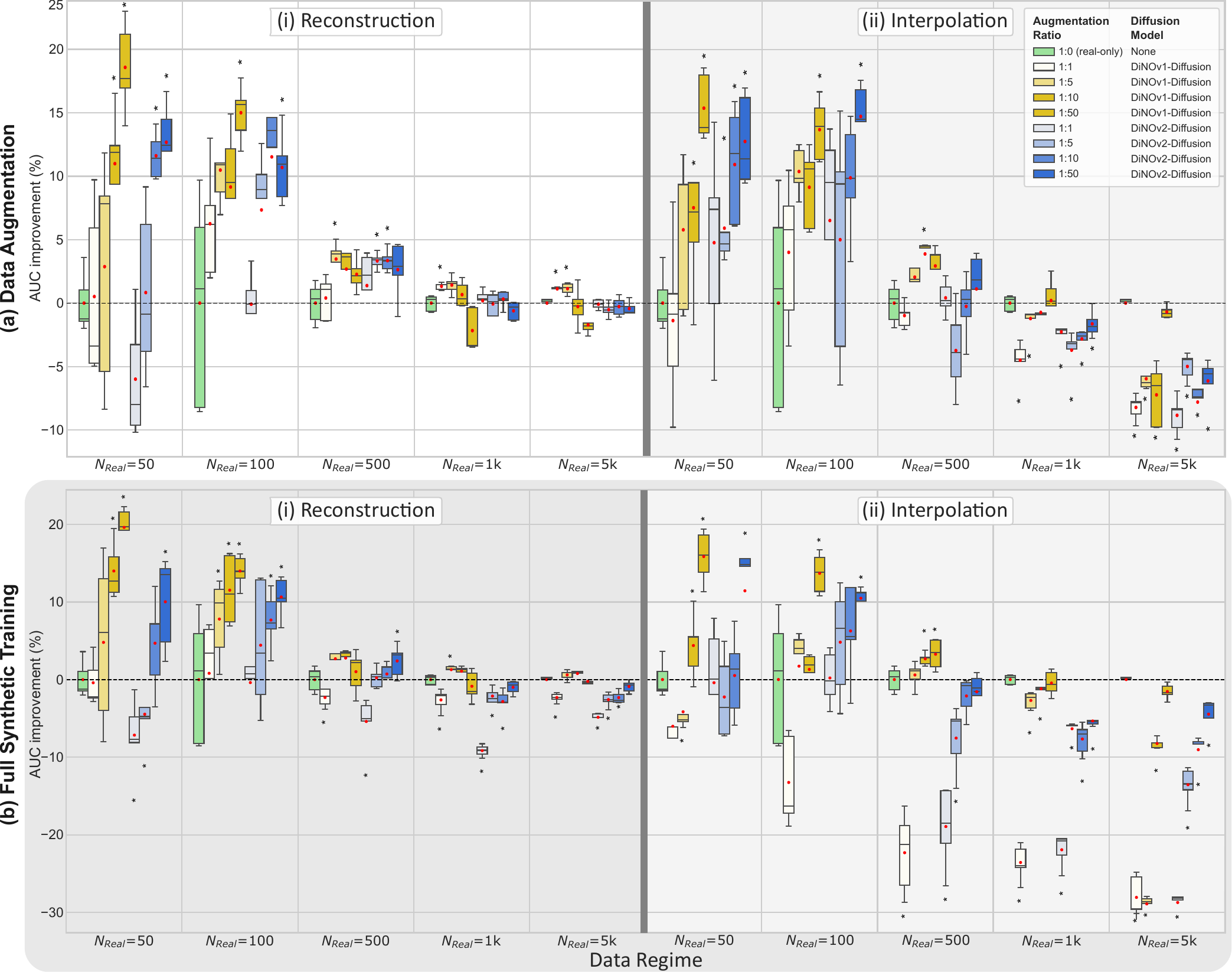}
    \caption{\footnotesize Boxplots for the Data Augmentation (a) and the Full Synthetic Training (b) experiments, representing performance improvement when adding synthetic data in different data regimes relative to using real data only. The horizontal line represents a 0\% improvement over the mean (red dot) classification performance when using real-data only (green bars) for each data regime and Real-to-Synthetic ($rs$) ratio independently. Therefore, values above the dotted line represent performance improvement and values below, performance degradation. The vertical lines separate the different data regimes for easier comparison, where the performance of DiNOv1-Diffusion (yellow palette) and DiNOv2-Diffusion (blue palette) are jointly displayed. In (i), the results for the reconstruction experiment are explored, whereas (ii) depicts the results for the interpolation experiment. Asterisks (*) represent statistical significance relative to real baseline ($p < 0.05$).}
    \label{fig:results-classification-experiments}
\end{figure*}

\subsection{Image Quality \& Checkpoint Selection}
\label{sec:results-image-quality}

The FID scores were calculated every 2500 steps over a subset of MIMIC's p19 dataset following \cite{chambon2022roentgen, mimic}. Both the DiNOv1 and DiNOv2 models converged relatively late, reaching scores of 4.7 and 6.4 at 80k and 120k steps, respectively. The full FID scores for every checkpoint can be observed in Figure~\ref{fig:imagequalitymetrics}. DiNOv1-Diffusion leads to lower FID scores when compared to DiNOv2-Diffusion. This is also evident by a slightly less saturated synthetic images generated with DiNOv2-Diffusion when compared to the source real images (see Figure \ref{fig:results-generated-images}).

\subsection{Data Augmentation}
\label{sec:results-data-augmentation}

In this experiment, real and synthetic data were used in different proportions to train DenseNet-121 classification models. Table~\ref{tab:results-classification-experiments}-a and Figure~\ref{fig:results-classification-experiments}-a provide the results of the cross-validation trainings. The 'reconstruction' workstream (see Section~\ref{sec:methodology-generation-reconstruction}) depicts consistent improvements when used for data augmentation in all data regimes, with AUC increases up to approximately 20\% in small-data regimes. In some larger-data regimes ($N \in [1000, 5000]$), the addition of large amounts of synthetic data slightly degraded performance, although never by a significant margin ($p > 0.05$). The 'interpolation' workstream (see Section~\ref{sec:methodology-generation-interpolation}) also depicts improvements in smaller data regimes as compared to not using synthetic data, although it leads to a significant performance degradation in large-data regimes ($p < 0.05$). Also, DiNO-Diffusion using DiNOv1 yields larger performance improvements compared to when using DiNOv2. This is always true for both image synthesis strategies, except for the interpolation results on data regime $N_{real}=100$, where the best test AUC is achieved with DiNOv2 for 1:50 $rs$ ratio.

\subsection{Full Synthetic Training}
\label{sec:results-cas}

The test set results of the full synthetic trainings are shown in Table~\ref{tab:results-classification-experiments}-b and Figure \ref{fig:results-classification-experiments}-b. The data synthesised via the ``reconstruction'' strategy (see Section~\ref{sec:methodology-generation-reconstruction}) using DiNOv1-Diffusion provided good performance in almost all settings, where statistically significant performance decreases only existed for the lowest $rs$ ratio in the largest three data regimes. For both ``reconstruction'' DiNO-Diffusion variants, training with sufficiently large $rs$ ratios in small-data regimes ($N_{real} \in [50, 100, 500]$) led to significant performance improvements of up to 20\%, mirroring the data augmentation results (see Section \ref{sec:methods-evaluation-data-augmentation}). However, for the ``interpolation'' based synthesis (see Section ~\ref{sec:methodology-generation-interpolation}), this was only the case in the 1:50 ratio.  Generally, the data synthesised via the ``interpolation'' strategy did not reliably train the classifier in splits larger than $N_{real} = 1k$ for DiNOv1-Diffusion and $N_{real} = 500$ for DiNOv2-Diffusion. Finally, DiNOv1-Diffusion yielded larger performance improvements and statistical significance when compared to DiNOv2-Diffusion.

\subsection{Zero-Shot Segmentation}
\label{sec:results-segmentation}

\begin{figure}[t]
    \centering
    \includegraphics[width=\linewidth]{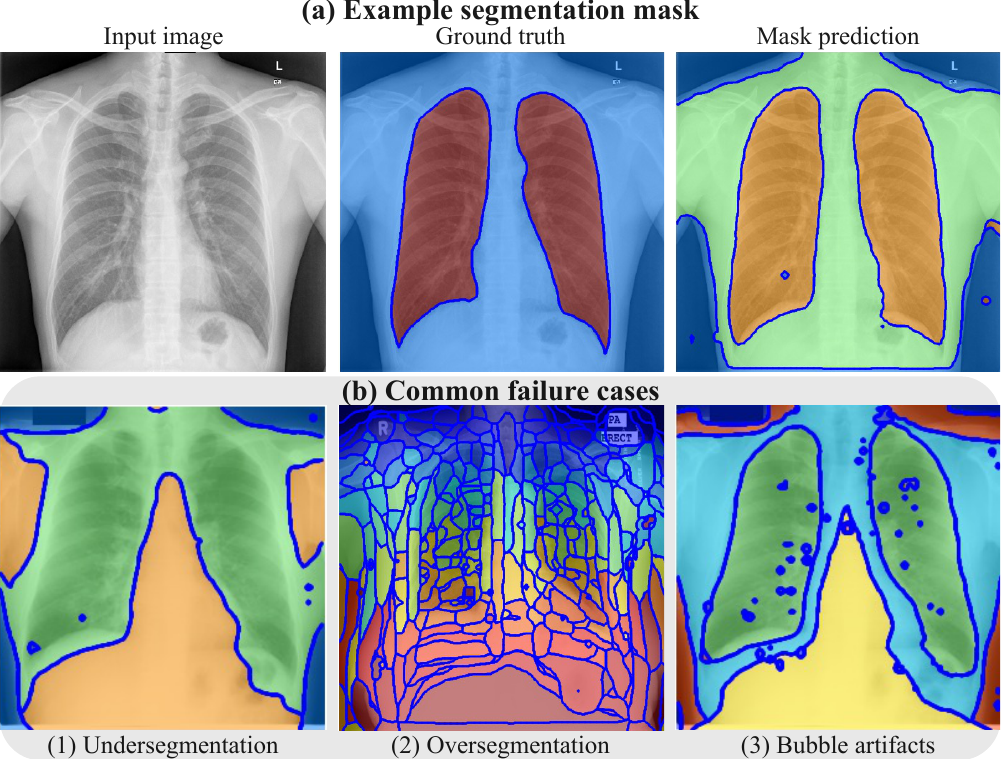}
    \caption{(a) Example segmentation masks generated by the best DiNOv1-Diffusion model and (b) common failure cases. Failures are caused by sub-optimal hyperparameters: (1) incomplete segmentation, often observed in early checkpoints or high thresholds; (2) oversegmentation and fragmentation, usually due to low merge thresholds; (3) bubble-like artifacts, mostly observed in later checkpoints.}
    \label{fig:segmentation_sample_results}
\end{figure}

The performance of the zero-shot experiments are shown in  Table~\ref{tab:segmentation_results}. Both DiNOv1- and DiNOv2-Diffusion showed improvements of up to 10\% Dice score when compared to a vanilla SD v1.5 model while also presenting lower variance. When addressing individual results, DiNOv1-Diffusion generated the best average Dice scores. Performance varied between datasets, with the Montgomery \cite{montgomeryshenzen} producing the lowest Dice scores for both vanilla Stable Diffusion and DiNOv1-Diffusion, but to the highest scores for the DiNOv2-based approach when comparing the overall best model. It should be noted that the best model checkpoint for segmentation was significantly earlier than the one found in Section~\ref{sec:results-image-quality}. 
Moreover, the optimal parameters for DiffSeg were very similar for both self-supervised DMs, while the optimal merging threshold was 10x larger for the base SD model. Finally, non-optimal combinations of parameters produced significant artifacts in the generated masks as shown in Figure~\ref{fig:segmentation_sample_results} (b).

\section{Discussion}
\label{sec:discussion}

DMs are a cornerstone in modern foundation models, revolutionizing many tasks in Computer Vision. Their ability to generate high-quality images has caused a large scientific, economic and societal disruption, whose long-term repercussions are difficult to foresee \cite{liu2024sora}. However, despite their scientific and industrial utility, applying this technology in medical imaging is severely limited by key challenges such as a lack of large-scale labeled datasets including high-quality textual or non-textual descriptions \cite{kazerouni2023medicaldiffusionsurvey}. Although this limitation might be temporary due to current trends in AI data acquisition and improved dataset interoperability \cite{mubashara2024croissant}, it is not clear whether the prevalent text-to-image generative recipe \cite{rombach2022stablediffusion} is optimal for medical applications.

\begin{table}[t]
\centering
\renewcommand{\arraystretch}{1.2}
\begin{adjustbox}{width=1\linewidth}
\begin{tabular}{|c|c|c|c||c|}
    \hline
    \rowcolor[gray]{.85}
    \begin{tabular}[c]{@{}c@{}}\\ \textbf{Dataset}\end{tabular} & \begin{tabular}[c]{@{}c@{}}\textbf{Stable} \\ \textbf{Diffusion 1.5}\end{tabular} & \begin{tabular}[c]{@{}c@{}} \textbf{DiNOv1-} \\ \textbf{\textbf{Diffusion}}\end{tabular}   & \begin{tabular}[c]{@{}c@{}} \textbf{DiNOv2-} \\ \textbf{\textbf{Diffusion}}\end{tabular}    &       \begin{tabular}[c]{@{}c@{}}\\ \textbf{\textbf{Supervised}}\end{tabular}  \\
    \hline
    Threshold  & 0.5   & 0.05             & 0.05          & - \\
    Timestep   & 300  & 300           & 300         & - \\
    Grid size  & 32x32 & 16x16& 16x16 & -\\
    \hline
    Shenzhen   & $80.7\pm15.9$ & $\mathbf{84.2\pm10.5}$ & $82.3\pm15.7$          & $98.3$ \cite{xu23lung}\\
    JSRT       & $80.9\pm12.1$ & $\mathbf{88.4\pm6.8}$ & $84.7\pm11.3$          & $97.9$\cite{liu2022automatic} \\
    Montgomery & $77.3\pm8.8$ & $78.3\pm8.6$          & $\mathbf{87.1\pm3.4}$ & $97.7$\cite{liu2022automatic} \\
    \hline
    \rowcolor[RGB]{223,230,245} Combined   & $80.3\pm14.2$ & $\mathbf{84.4\pm9.9}$ & $83.6\pm13.6$       & -   \\
    \hline
\end{tabular}
\end{adjustbox}
\caption{\footnotesize Segmentation performance, measured by mean Dice scores (\%). The displayed values are based on the hyperparameter configurations that led to best overall results.}
\label{tab:segmentation_results}
\end{table}

Some approaches employing DMs in medical data exist. Chambon \textit{et al.} \cite{chambon2022roentgen} trained an SD architecture on the MIMIC-CXR dataset \cite{mimic} with good synthesis fidelity, reporting low FID scores and high accuracy scores on several downstream tasks including classification, report generation and image retrieval. However, their approach is severely limited on the size of the development dataset (300k images) and the low quality of accompanying captions. In histopathology, multiple authors have proposed applying DMs for image generation \cite{ye2023histodiffusion, aversa2023diffinfinite, xu2024vitdae, osorio2023latent}. For instance, Aversa \textit{et al}. relied on a custom-annotated dataset of large histopathology slides with segmentation masks representing different tissue subtypes within the slide and employed timestep unravelling to generate images larger than the typical 512$\textrm{px}^2$.
However, their approach heavily relied on a closed-source, custom-annotated dataset, and timestep unraveling might be impractical in other medical imaging modalities.
In contrast, Xu \textit{et al}. \cite{xu2024vitdae} take a similar approach as the one proposed here, and train a DM conditioned only on an image encoder's $c_{CLS}$ for histopathology image synthesis. However, their method was partially supervised, as it relied on training additional label-specific DMs for $c_{CLS}$ generation. Besides being compute intensive, their method fails to leverage the emerging data augmentation and segmentation capabilities that a self-supervision DM training conveys. Finally, Pinaya \textit{et al.} \cite{pinaya2022brainmridiffusion} trained an LDM on a large dataset of 31740 3D Brain MRI images from UK BioBank. However, despite the scale of this dataset, the fragmentation of clinical labels forced the authors to condition the DM with simplified clinical variables such as age, sex, ventricular volume, and brain volume.

DiNO-Diffusion addresses the data limitations in medical imaging by conditioning the image generation process on the images themselves. This allows training DMs on unlabelled data, which is more abundant in the medical field. The resulting DiNO-Diffusion models demonstrated good manifold coverage, as indicated by low FID scores, and exhibited notable properties in several downstream tasks. Firstly, adding synthetic data using the ``reconstruction'' strategy improved performance across most configurations. However, performance gains diminished as more real data became available, which is to be expected. Secondly, the ``interpolation'' strategy degraded performance in higher data regimes. We hypothesize that, although the generated images qualitatively resemble plausible images (see Figure~\ref{fig:results-generated-images}-b), naïvely interpolating embeddings did not ensure that the interpolated labels corresponded to the decoded image's features, thereby hurting classification performance. We leave to future work the exploration of more sophisticated interpolation strategies. 
Thirdly, full synthetic training demonstrated that synthetic data can replace real data while preserving privacy, and even improve performance in small-data regimes, when used in abundance. Finally, DiNO-Diffusion's zero-shot segmentation outperformed a vanilla SD architecture. This is remarkable given that the dataset used to train the vanilla SD model was several orders of magnitude larger. 

Despite DiNO-Diffusion's performance, conditioning the synthesis process on image embeddings has theoretical advantages and disadvantages. This type of conditioning relaxes the need for annotations, enabling the collection of larger datasets for model training, and has proven effective across various tasks. However, usage of an image-conditioned model is fundamentally different from text-based approaches, as image generation requires conditioning on an image. Still, this circular dependency between input and output could be advantageous in some use cases, such as data augmentation or privacy-preserving data sharing. 

These advantages and disadvantages evidence room for improvement. Firstly, DiNOv1-Diffusion outperformed DiNOv2-Diffusion both quantitatively and qualitatively, despite the larger data pool used to train the DiNOv2 image encoder \cite{oquab2023dinov2}. This suggests that using domain-specific encoders \cite{cohen2022torchxrayvision, perez2024raddino, moutakanni2024advancingraydino}, or even a combination of different image encoders \cite{esser2024stablediffusion3, liu2024sora} could further improve these results.
Secondly, DiNO-Diffusion would benefit from more recent diffusion architectures found in the literature \cite{esser2024stablediffusion3, liu2024sora, betker2023dalle3}.
Thirdly, generation based on other descriptors, such as text, could be enabled by bridging the gap to the image embedding space, e.g., via lean mapping networks \cite{zhang2023controlnet, li2023blip2} or by training other conditional generative models \cite{xu2024vitdae}.
Finally, the failure cases found in the zero-shot segmentation workstream require adapting the DiffSeg methodology to datasets with different characteristics. These adaptations could include image-level hyperparameter optimization, more sophisticated attention-merging strategies, or using DiNO's attention maps to better locate anatomic structures.

\section{Conclusions}
\label{sec:conclusions}

Diffusion models have significantly impacted the Computer Vision community, offering unprecedented capabilities in high-quality image generation with broad scientific, economic, and societal implications. However, their application to medical imaging is constrained by data and annotation scarcity. Our DiNO-Diffusion approach addresses this problem by conditioning the image generation on the images themselves, eliminating the need for extensive annotations. The approach shows promising results in manifold coverage, data augmentation, privacy preservation and zero-shot segmentation. The image-conditioned nature of the model changes the inference workflow. This is likely beneficial in some use cases, and promising paths exist to address limitations in others. This work highlights the efficacy of DiNO-Diffusion in medical imaging and underscores the need for innovative domain-specific solutions to fully leverage the potential of diffusion models in medical imaging.


\end{document}